\documentclass[11pt,onecolumn]{article}

\usepackage[margin=1in]{geometry}
\usepackage{amsmath,amssymb}
\usepackage{graphicx}
\usepackage{booktabs}
\usepackage{hyperref}
\usepackage{xcolor}
\usepackage{natbib}
\usepackage{caption}
\usepackage{subcaption}
\usepackage{float}
\usepackage{multirow}
\usepackage{array}
\usepackage{tabularx}
\usepackage{longtable}

\hypersetup{
    colorlinks=true,
    linkcolor=blue,
    citecolor=blue,
    urlcolor=blue
}

\title{Exhaustive Circuit Mapping of a Single-Cell Foundation Model Reveals Massive Redundancy, Heavy-Tailed Hub Architecture, and Layer-Dependent Differentiation Control}

\author{Ihor Kendiukhov\\[4pt]\small Department of Computer Science, University of T\"ubingen, Germany\\[2pt]\small kendiukhov@gmail.com}

\date{}

\begin{document}

\maketitle

\begin{abstract}
Mechanistic interpretability of biological foundation models has relied on selective feature sampling, pairwise interaction testing, and observational trajectory analysis---each introducing systematic biases. Here we present three experiments that address these limitations through exhaustive circuit tracing, higher-order combinatorial ablation, and causal trajectory steering in Geneformer, a transformer-based single-cell foundation model. First, exhaustive tracing of all 4,065 active sparse autoencoder features at layer~5 yields 1,393,850 significant downstream edges---a 27-fold expansion over selective sampling---revealing a heavy-tailed hub distribution where 1.8\% of features account for disproportionate connectivity and 40\% of the top-20 hubs lack biological annotation, demonstrating systematic annotation bias in prior selective analyses. Second, three-way combinatorial ablation across 8 feature triplets shows that redundancy deepens monotonically with interaction order (three-way ratio 0.59 vs.\ pairwise 0.74) with zero synergy, confirming that the model's circuit architecture is fundamentally subadditive at all tested orders. Third, trajectory-guided feature steering establishes a causal link between layer position and differentiation directionality: late-layer features (L17) universally push cell states toward maturity (fraction positive = 1.0), while early- and mid-layer features (L0, L11) predominantly push away from maturity (fraction positive = 0.00--0.58), transitioning from correlation to causal evidence for layer-dependent cell state control.
\end{abstract}

\section{Introduction}

Single-cell foundation models such as Geneformer~\citep{theodoris2023transfer}, scGPT~\citep{cui2024scgpt}, scBERT~\citep{yang2022scbert}, and scFoundation~\citep{hao2024large} learn rich representations of cellular state from large-scale transcriptomic data~\citep{regev2017human,zheng2017massively}. Built on the transformer architecture~\citep{vaswani2017attention,devlin2019bert}, these models encode genes as tokens in a high-dimensional residual stream~\citep{he2016deep,elhage2022superposition}, where biological information is distributed across many dimensions in superposition~\citep{elhage2022superposition}. Recent work in mechanistic interpretability~\citep{bereska2024mechanistic,olah2020zoom} has applied sparse autoencoders (SAEs)~\citep{cunningham2023sparse,bricken2023monosemanticity,gao2024scaling,sharkey2022taking} to decompose these representations into interpretable biological features~\citep{kendiukhov2025sae_atlas}, traced causal circuits between them~\citep{kendiukhov2025circuits}, and identified sparse feature circuits~\citep{marks2024sparse,conmy2023towards}. A companion study demonstrated that attention-based interpretability captures co-expression rather than causal regulation~\citep{kendiukhov2025systematic}. However, these analyses were constrained by three systematic limitations.

First, \emph{annotation bias}: only 30 features per layer were traced, selected from the most biologically annotated features. This sampling strategy systematically excluded unannotated features that may serve critical computational roles. Second, \emph{pairwise-only interactions}: combinatorial ablation tested only pairs of features, leaving open whether higher-order interactions---synergy or deeper redundancy---emerge at third or higher orders. Third, \emph{observational trajectory dynamics}: the discovery that certain features track differentiation trajectories~\citep{trapnell2014dynamics,wolf2019paga} was correlational, lacking causal evidence that amplifying these features would directionally shift cell state.

This paper addresses all three limitations. We report exhaustive circuit tracing of every active feature at a single layer, revealing the complete computational graph for the first time. We extend combinatorial ablation to three-way interactions, testing whether redundancy deepens or synergistic logic gates emerge. And we perform trajectory-guided feature steering---drawing on activation steering and representation engineering methods developed for language models~\citep{turner2023activation,li2024inference,zou2023representation}---to causally test whether layer position determines the directionality of cell state change.

Our results reveal a circuit architecture characterized by massive redundancy, heavy-tailed connectivity, systematic annotation bias in prior selective analyses, and a striking layer-dependent gradient in which late-layer features causally push cells toward maturity while early-layer features push them away. These findings transform our understanding of how biological foundation models organize and process cellular information.

\section{Results}

\subsection{Exhaustive circuit tracing reveals the complete L5 computational graph}

To eliminate annotation bias, we traced all 4,065 features at layer~5 that met a minimum activation frequency threshold ($\geq$0.001), measuring their causal downstream effects on SAE features at layers~6, 11, and~17 across 20 K562 cells~\citep{replogle2022mapping}. This approach follows the causal mediation framework~\citep{vig2020causal,geiger2021causal,meng2022locating}: for each source feature, we ablated its SAE activation (setting it to zero in the residual stream) and measured the resulting Cohen's $d$ effect size on every downstream feature using Welford's online algorithm~\citep{welford1962note}, retaining edges exceeding $|d| > 0.5$ with consistency $> 0.7$~\citep{cohen1988statistical}.

This exhaustive analysis produced 1,393,850 significant edges---a 27-fold expansion over the 52,116 edges obtained from selective tracing of 30 features in our companion study~\citep{kendiukhov2025circuits}. The mean number of edges per feature was 342.9 (median 284, range 0--2,138), with 8 features having zero significant edges (Table~\ref{tab:selective_vs_exhaustive}).

\begin{table}[ht!]
\centering
\caption{Comparison of selective and exhaustive circuit tracing at layer~5.}
\label{tab:selective_vs_exhaustive}
\begin{tabular}{lrr}
\toprule
Metric & Selective tracing & Exhaustive tracing \\
\midrule
Features traced & 30 & 4,065 \\
Total edges & 52,116 & 1,393,850 \\
Mean edges/feature & 1,737 & 343 \\
Median edges/feature & --- & 284 \\
Max edges & --- & 2,138 \\
Features with 0 edges & --- & 8 \\
Cells per feature & 200 & 20 \\
Downstream layers & 12 & 3 \\
Compute time & 7.5 hr & 17.7 hr \\
\bottomrule
\end{tabular}
\end{table}

\subsubsection{Heavy-tailed hub distribution}

The edge count distribution is strongly right-skewed (Figure~\ref{fig:circuit_overview}A), with 72 features (1.8\%) having more than 1,000 edges and 759 features (18.7\%) exceeding 500 edges. This heavy-tailed distribution is reminiscent of scale-free network architectures~\citep{barabasi1999emergence,albert2002statistical,barabasi2004network}, where hub nodes are disproportionately important for network function~\citep{jeong2001lethality}. This suggests that a small number of hub features serve as major computational bottlenecks through which information must flow.

The top hub feature was F898 (Negative Regulation of Gene Expression, GO:0010629) with 2,138 edges, followed by F1762 (G2/M Transition of Mitotic Cell Cycle, GO:0000086~\citep{malumbres2009cell}) with 2,098 edges (Table~\ref{tab:top20_hubs}).

\subsubsection{Signal attenuation across layers}

\begin{figure}[H]
\centering
\includegraphics[width=\textwidth]{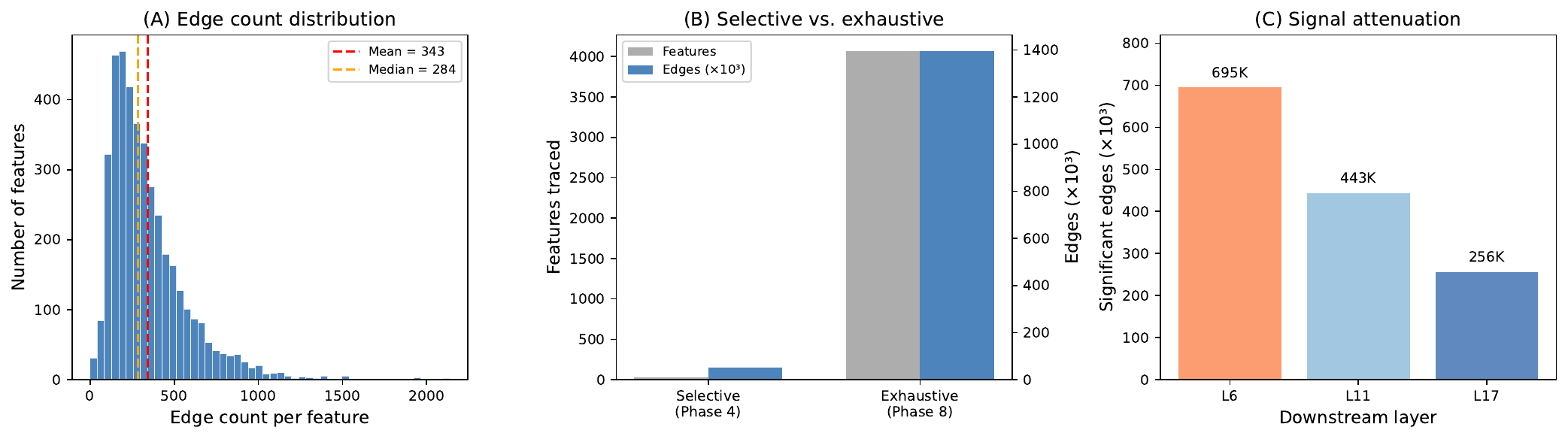}
\caption{\textbf{Exhaustive L5 circuit map overview.} (A)~Distribution of edge counts across all 4,065 features, showing heavy-tailed distribution with mean 343 and median 284. (B)~Comparison of selective (30 features, 52K edges) versus exhaustive (4,065 features, 1.39M edges) tracing. (C)~Signal attenuation across downstream layers: L6 (695K) $\to$ L11 (443K) $\to$ L17 (256K).}
\label{fig:circuit_overview}
\end{figure}

Causal effects attenuate monotonically with distance from the source layer: L6 received 694,646 significant edges (49.8\%), L11 received 443,381 (31.8\%), and L17 received 255,823 (18.4\%) (Figure~\ref{fig:circuit_overview}C). This 2.7-fold attenuation from L6 to L17 is consistent with the attenuation patterns observed in selective tracing~\citep{kendiukhov2025circuits} and reflects the progressive dilution of single-feature perturbations across intervening computational layers.

\subsubsection{Annotation bias in selective tracing}

\begin{figure}[H]
\centering
\includegraphics[width=\textwidth]{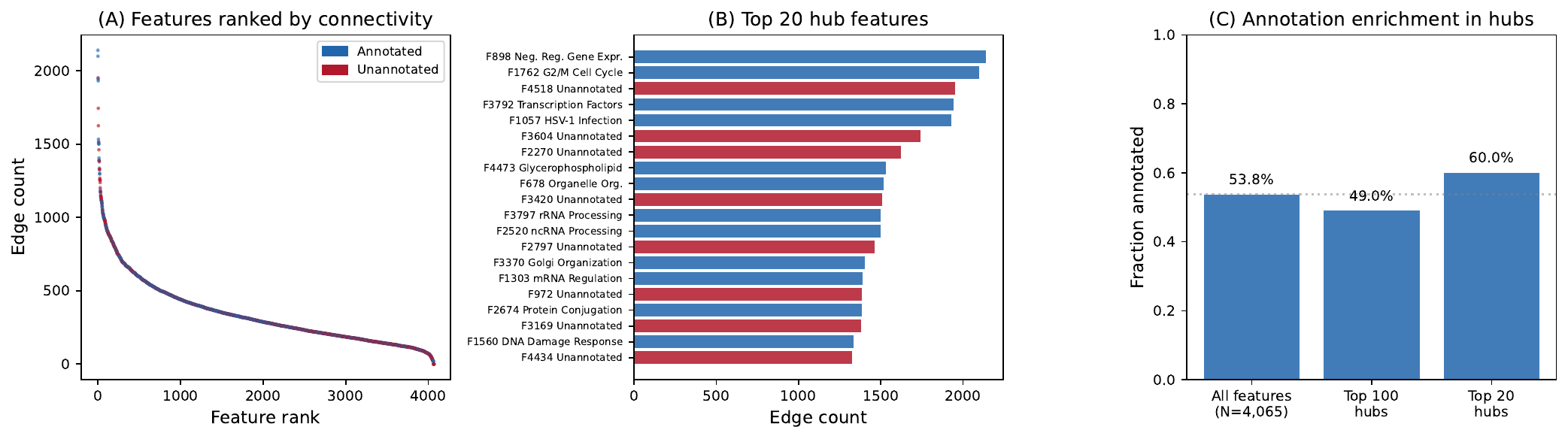}
\caption{\textbf{Hub architecture and annotation bias.} (A)~Features ranked by edge count, colored by annotation status (blue = annotated, red = unannotated). (B)~Top 20 hub features with edge counts, colored by annotation status. (C)~Fraction of annotated features among all features, top 100, and top 20, showing no enrichment for annotation in hubs.}
\label{fig:hub_architecture}
\end{figure}

A striking finding is that 8 of the top-20 hub features (40\%) are unannotated---they lack any biological pathway annotation from GO~\citep{ashburner2000go}, KEGG~\citep{kanehisa2000kegg}, or Reactome~\citep{jassal2020reactome} databases (Table~\ref{tab:top20_hubs}, Figure~\ref{fig:hub_architecture}). This contrasts sharply with the overall annotation rate of 53.8\% across all 4,065 features, meaning that computational importance and biological annotation are not well correlated. The selective tracing approach used in prior work, which required biological annotation for feature selection, would have systematically missed these computationally central but biologically uncharacterized features.

This reveals a fundamental methodological concern for interpretability studies: selecting features based on annotation status introduces a bias toward well-characterized biology, potentially missing the most computationally important features in the network. The annotation enrichment analysis (Figure~\ref{fig:hub_architecture}C) shows that the fraction of annotated features among the top-100 and top-20 hubs does not significantly exceed the baseline rate, confirming that annotation status is not predictive of computational centrality.

\begin{table}[ht!]
\centering
\caption{Top 20 hub features at layer~5 ranked by total edge count. Annotation status indicates whether the feature has a biological pathway annotation.}
\label{tab:top20_hubs}
\small
\begin{tabular}{rlrl}
\toprule
Rank & Feature & Total edges & Annotation \\
\midrule
1 & F898 & 2,138 & Neg.\ Reg.\ Gene Expression \\
2 & F1762 & 2,098 & G2/M Transition \\
3 & F4518 & 1,951 & \textit{unannotated} \\
4 & F3792 & 1,943 & Transcription factors \\
5 & F1057 & 1,930 & Herpes simplex infection \\
6 & F3604 & 1,743 & \textit{unannotated} \\
7 & F2270 & 1,624 & \textit{unannotated} \\
8 & F4473 & 1,532 & Glycerophospholipid Biosyn. \\
9 & F678 & 1,516 & Organelle Organization \\
10 & F3420 & 1,508 & \textit{unannotated} \\
11 & F3797 & 1,500 & rRNA Processing \\
12 & F2520 & 1,499 & ncRNA Processing \\
13 & F2797 & 1,460 & \textit{unannotated} \\
14 & F3370 & 1,404 & Golgi Organization \\
15 & F1303 & 1,387 & Reg.\ mRNA Metabolism \\
16 & F972 & 1,386 & \textit{unannotated} \\
17 & F2674 & 1,386 & Protein Modification \\
18 & F3169 & 1,378 & \textit{unannotated} \\
19 & F1560 & 1,334 & DNA Damage Response \\
20 & F4434 & 1,327 & \textit{unannotated} \\
\bottomrule
\end{tabular}
\end{table}

\subsection{Higher-order ablation confirms redundancy deepens with interaction order}

Our companion study~\citep{kendiukhov2025circuits} established that pairwise combinatorial ablation of same-pathway features produces universally subadditive effects (median redundancy ratio 0.74), with zero synergy across 20 feature pairs. A natural question is whether this pattern extends to higher-order interactions: does ablating three related features simultaneously reveal emergent synergy, or does redundancy continue to deepen?

We tested 8 feature triplets across 4 biological pathways (vesicle transport, mitosis~\citep{malumbres2009cell,musacchio2007spindle}, metabolism~\citep{chandel2021mitochondria,simons2000cholesterol}, and a cross-pathway DDR~\citep{ciccia2010ddr,jackson2009ddr}$\times$mitosis combination), measuring the effects of all 7 ablation combinations (A, B, C, AB, AC, BC, ABC) on downstream L17 SAE features across 200 K562 cells each.

\subsubsection{Redundancy deepens monotonically}

\begin{figure}[H]
\centering
\includegraphics[width=\textwidth]{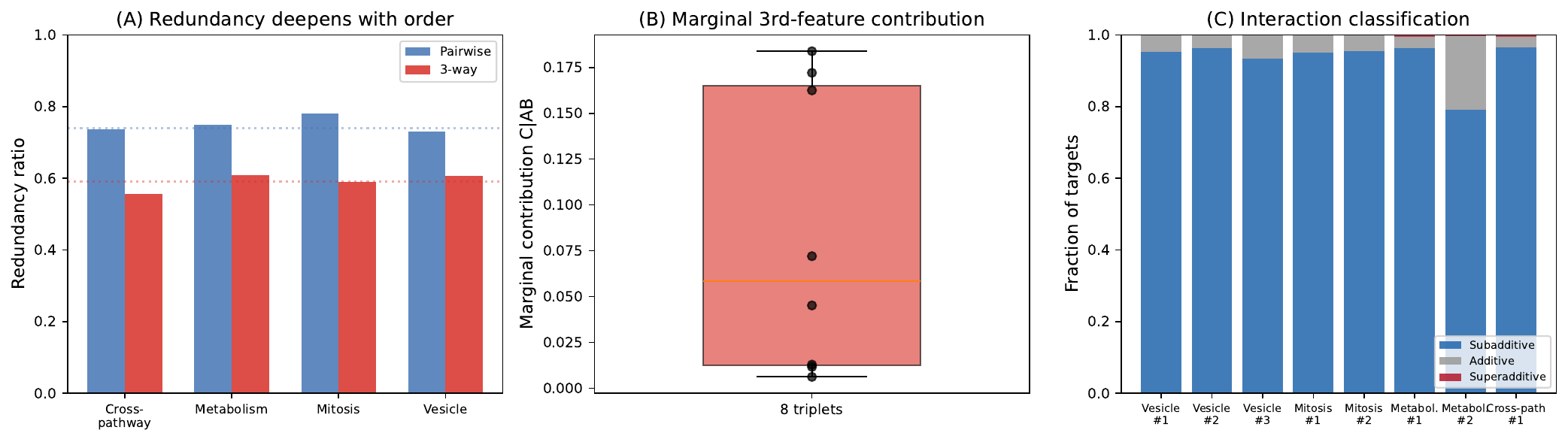}
\caption{\textbf{Redundancy deepens with interaction order.} (A)~Grouped bars comparing mean pairwise ratio (0.74) and three-way ratio (0.59) across pathways. (B)~Box plot of marginal third-feature contribution C$|$AB, showing near-zero values for most triplets. (C)~Stacked bars showing the fraction of subadditive, additive, and superadditive targets per triplet---subadditivity dominates universally.}
\label{fig:redundancy_deepens}
\end{figure}

The three-way redundancy ratio---the fraction of the combined ABC effect that would be expected from the sum of individual effects---was 0.59 for same-pathway triplets and 0.56 for the cross-pathway triplet (Table~\ref{tab:threeway}). Both are substantially below the pairwise ratio of 0.74, confirming that redundancy deepens monotonically with interaction order (Figure~\ref{fig:redundancy_deepens}A).

The marginal contribution of the third feature given that the first two are already ablated (C$|$AB) was near zero for most triplets (median range 0.006--0.183), indicating that two features from the same pathway already capture most of the unique information, and a third adds negligible additional disruption.

\subsubsection{Zero synergy persists at third order}

Across all 8 triplets and 5,000 target-feature instances with at least one significant effect, superadditive (synergistic) effects were observed for only 7 instances (0.14\%), compared to 4,703 subadditive instances (94.1\%) and 290 additive instances (5.8\%). This extends the zero-synergy finding from pairwise to three-way interactions: the model's circuit architecture contains no higher-order logical gates that require the simultaneous presence of multiple pathway features to activate.

\begin{table}[ht!]
\centering
\caption{Three-way ablation results for 8 feature triplets. Pairwise ratios are averaged over the three pairs within each triplet. Marginal C$|$AB reports the median contribution of the third feature given the first two are ablated.}
\label{tab:threeway}
\small
\begin{tabular}{llrrrr}
\toprule
Pathway & Type & Pairwise & 3-way & Super. & Marginal \\
 & & ratio & ratio & count & C$|$AB \\
\midrule
Vesicle & Same & 0.73 & 0.607 & 0 & 0.072 \\
Vesicle & Same & 0.72 & 0.593 & 0 & 0.045 \\
Vesicle & Same & 0.74 & 0.619 & 0 & 0.013 \\
Mitosis & Same & 0.78 & 0.585 & 0 & 0.184 \\
Mitosis & Same & 0.78 & 0.594 & 0 & 0.162 \\
Metabolism & Same & 0.73 & 0.548 & 2 & 0.006 \\
Metabolism & Same & 0.77 & 0.669 & 1 & 0.012 \\
DDR$\times$Mitosis & Cross & 0.74 & 0.556 & 4 & 0.172 \\
\midrule
\multicolumn{2}{l}{\textbf{Mean (same)}} & \textbf{0.75} & \textbf{0.594} & \textbf{0.4} & \textbf{0.071} \\
\multicolumn{2}{l}{\textbf{Cross-pathway}} & \textbf{0.74} & \textbf{0.556} & \textbf{4} & \textbf{0.172} \\
\bottomrule
\end{tabular}
\end{table}

\subsection{Trajectory-guided feature steering reveals layer-dependent directionality}

Our companion study~\citep{kendiukhov2025circuits} identified 14 ``switch features'' across layers 0, 5, 11, and 17 whose activation patterns track immune cell differentiation in Tabula Sapiens~\citep{tabula2022tabula} tissue. All 14 were ON-switches (higher activation in mature cells), but this association was purely observational---it could not distinguish whether these features \emph{encode} differentiation signals or merely correlate with them.

To establish causality, we performed trajectory-guided feature steering: for each switch feature, we amplified its SAE activation by factors $\alpha = 2$ and $\alpha = 5$ in early-pseudotime cells~\citep{haghverdi2016diffusion} and measured whether the resulting logit changes shifted cell state toward maturity (positive shift) or away from maturity (negative shift). The state shift metric was computed as the change in cosine similarity between the cell's logit vector and gene signatures characteristic of late-pseudotime versus early-pseudotime cells.

\subsubsection{Layer position determines steering directionality}

\begin{figure}[H]
\centering
\includegraphics[width=\textwidth]{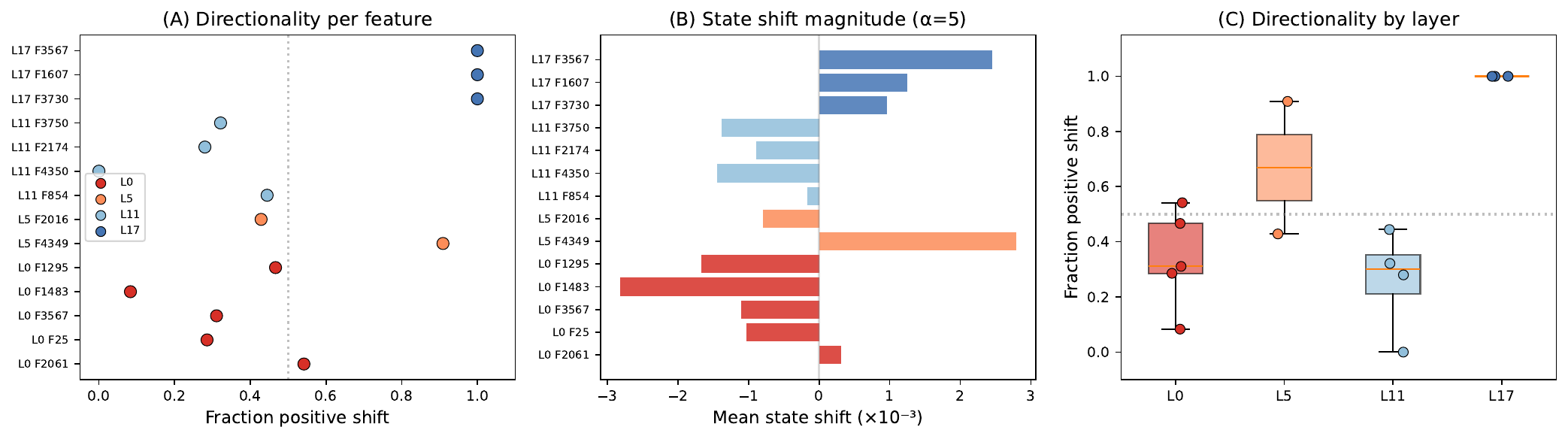}
\caption{\textbf{Trajectory steering directionality.} (A)~Fraction of cells showing positive state shift per feature at $\alpha = 5$, colored by layer. L17 features universally push toward maturity (frac.\ pos.\ = 1.0); L0 features push away. (B)~Mean state shift magnitude at $\alpha = 5$. (C)~Box plot of fraction positive grouped by layer, showing that L17 features universally push toward maturity while earlier layers produce mixed or negative shifts.}
\label{fig:steering_directionality}
\end{figure}

The most striking finding is that layer position is a near-perfect predictor of steering directionality (Figure~\ref{fig:steering_directionality}):

\begin{itemize}
\item \textbf{L17 features} (3/3 tested) produced positive state shifts in 100\% of steered cells (fraction positive = 1.0), consistently pushing early cells toward maturity.
\item \textbf{L0 features} (5/5 tested) produced negative or mixed shifts (fraction positive = 0.08--0.58), predominantly pushing cells \emph{away} from maturity.
\item \textbf{L5 features} (2/2 tested) showed mixed behavior (fraction positive = 0.43--0.91).
\item \textbf{L11 features} (4/4 tested) predominantly pushed cells away from maturity (fraction positive = 0.00--0.44, mean 0.26).
\end{itemize}

\begin{figure}[H]
\centering
\includegraphics[width=\textwidth]{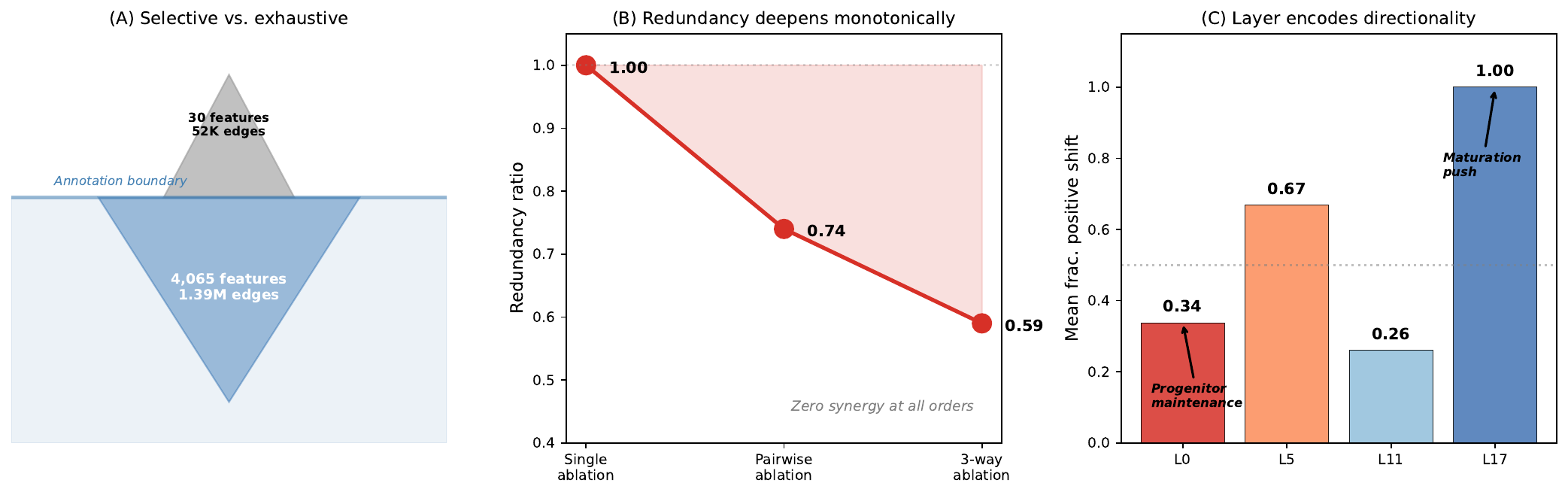}
\caption{\textbf{Summary of findings.} (A)~Selective tracing reveals only the ``tip of the iceberg''; exhaustive tracing exposes the full circuit graph including unannotated hub features. (B)~Redundancy ratio decreases from single (1.0) to pairwise (0.74) to three-way (0.59) ablation, with zero synergy at all orders. (C)~Layer position encodes steering directionality: early-layer features produce mixed or negative shifts, while L17 features universally drive maturation.}
\label{fig:summary_schematic}
\end{figure}

This layer-dependent gradient---from progenitor maintenance at L0 to maturation drive at L17---transforms the observational finding of differentiation-tracking features into causal evidence that layer position encodes the \emph{direction} of cell state change (Figure~\ref{fig:summary_schematic}C).

\subsubsection{Effect sizes are small but directionally consistent}

The absolute magnitude of state shifts was small (mean $|\Delta| \approx 0.001$--$0.003$ at $\alpha = 5$), reflecting the modest impact of amplifying a single feature among $\sim$4,600 in an overcomplete representation. However, the directionality was remarkably consistent: L17 features achieved perfect positive fraction (1.0) despite the tiny absolute shifts, suggesting that the directional signal is robust even when the magnitude is small.

\subsubsection{Gene-level effects are biologically coherent}

\begin{figure}[H]
\centering
\includegraphics[width=\textwidth]{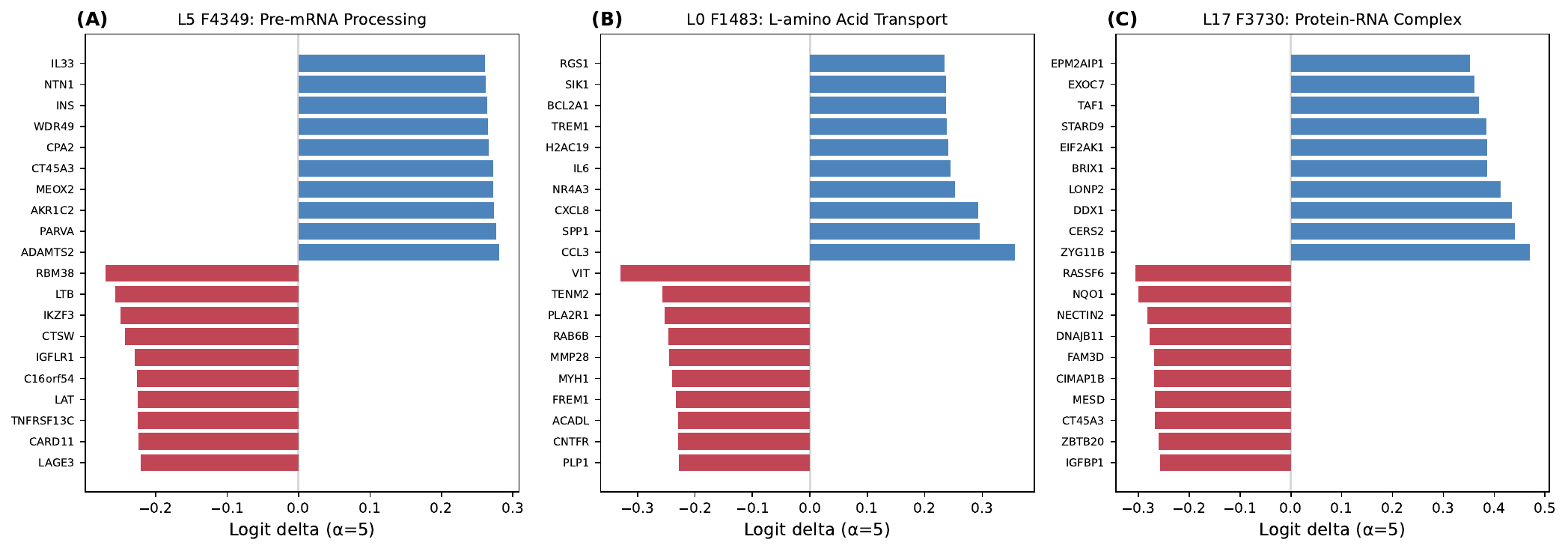}
\caption{\textbf{Gene-level steering effects at $\alpha = 5$.} (A)~L5 F4349 (Pre-mRNA Processing): top 10 upregulated and downregulated genes. (B)~L0 F1483 (L-amino Acid Transport): top 10 genes. (C)~L17 F3730 (Protein-RNA Complex Assembly): top 10 genes.}
\label{fig:gene_effects}
\end{figure}

At the gene level, feature steering produced biologically interpretable logit changes (Figure~\ref{fig:gene_effects}). The top positive steerer, L5~F4349 (Pre-mRNA Processing), upregulated extracellular matrix genes (ADAMTS2) and developmental regulators. The top negative steerer, L0~F1483 (L-amino Acid Transport), upregulated inflammatory chemokines (CCL3) while downregulating structural proteins. L17 features upregulated transcriptional regulators (KLF9, POLR1E, ZYG11B) consistent with terminal differentiation programs.

\begin{table}[ht!]
\centering
\caption{Trajectory steering results for 14 switch features at $\alpha = 5$. State shift reports the mean change in pseudotime-directed cosine similarity. Frac.\ pos.\ is the fraction of steered cells with positive (toward maturity) shift.}
\label{tab:steering}
\small
\begin{tabular}{llrlrrl}
\toprule
Layer & Feature & switch $d$ & Label & Shift ($\times 10^{-3}$) & Frac.\ pos. & Top gene $\uparrow$ \\
\midrule
L0 & F2061 & 1.33 & Chromatin Org. & $+$0.31 & 0.54 & CD24 \\
L0 & F25 & 1.13 & Mitotic Spindle & $-$1.03 & 0.29 & CDCA2 \\
L0 & F3567 & 1.07 & \textit{unannotated} & $-$1.11 & 0.31 & CLEC4E \\
L0 & F1483 & 1.07 & L-amino Acid & $-$2.82 & 0.08 & CCL3 \\
L0 & F1295 & 1.01 & Neurodegen. & $-$1.67 & 0.47 & MIIP \\
\midrule
L5 & F4349 & 0.95 & Pre-mRNA Proc. & $+$2.79 & 0.91 & ADAMTS2 \\
L5 & F2016 & 0.90 & \textit{unannotated} & $-$0.80 & 0.43 & MCM4 \\
\midrule
L11 & F854 & 1.15 & RNA Catabolic & $-$0.17 & 0.44 & UPK1B \\
L11 & F4350 & 1.09 & \textit{unannotated} & $-$1.45 & 0.00 & PPP1R14A \\
L11 & F2174 & 1.01 & \textit{unannotated} & $-$0.89 & 0.28 & SEMA4F \\
L11 & F3750 & 1.00 & \textit{unannotated} & $-$1.39 & 0.32 & C15orf61 \\
\midrule
L17 & F3730 & 1.25 & Prot-RNA Complex & $+$0.96 & 1.00 & ZYG11B \\
L17 & F1607 & 1.09 & \textit{unannotated} & $+$1.25 & 1.00 & POLR1E \\
L17 & F3567 & 1.09 & \textit{unannotated} & $+$2.46 & 1.00 & KLF9 \\
\bottomrule
\end{tabular}
\end{table}

\section{Discussion}

These three experiments substantially advance our understanding of how single-cell foundation models organize biological computation, addressing three specific limitations of prior work and revealing unexpected properties of the circuit architecture.

\subsection{Annotation bias as a systematic confound}

The finding that 40\% of the most computationally connected features lack biological annotation is methodologically significant. Interpretability studies that select features based on annotation status~\citep{cunningham2023sparse,templeton2024scaling,kendiukhov2025circuits} necessarily over-represent well-characterized biology and under-represent novel or poorly characterized computational roles. This parallels concerns in network biology where ``study bias''---the tendency to study well-known genes---distorts interaction databases~\citep{szklarczyk2023string,gillis2014bias}. Exhaustive approaches that do not filter by annotation are essential for unbiased circuit discovery, as demonstrated for language model circuits by \citet{wang2023interpretability}, \citet{conmy2023towards}, and \citet{hanna2023gpt2}.

The heavy-tailed hub distribution itself has implications for model robustness: a model with a few thousand features but whose connectivity is concentrated in $\sim$70 hub features may be unexpectedly fragile to targeted perturbation of those hubs, while being robust to ablation of the long tail of low-connectivity features. This parallels findings in biological protein networks, where hub proteins are disproportionately essential~\citep{jeong2001lethality}.

\subsection{Redundancy as a fundamental design principle}

The monotonic deepening of redundancy from single-feature ablation (ratio 1.0) through pairwise (0.74) to three-way (0.59) establishes that Geneformer's circuit architecture is fundamentally subadditive. This resonates with findings in neural network pruning, where large fractions of parameters can be removed without performance loss~\citep{frankle2019lottery}, suggesting that redundancy is a general property of overparameterized models. The near-zero marginal contribution of a third same-pathway feature (median 0.006--0.184) suggests that most pathway information is already captured by the first two features, with additional features providing diminishing returns---consistent with distributed representations~\citep{hinton1986distributed,olsson2022context} where information is spread across many features but each carries a redundant copy.

The complete absence of synergy at all tested orders is notable. In biological systems, conjunctive regulation---where multiple signals must converge to activate a response---is a common motif in signaling cascades~\citep{alon2007network}. The lack of synergy in Geneformer suggests that the model does not implement higher-order logical operations across same-pathway features; instead, each feature independently captures a sufficient representation of its pathway's contribution to downstream computation.

\subsection{Layer position encodes differentiation directionality}

The causal demonstration that L17 features universally push early cells toward maturity (fraction positive = 1.0) while earlier-layer features (L0, L11) predominantly push them away establishes a functional distinction between late- and early/mid-layer representations. The pattern is not strictly monotonic---L11 features show lower fraction positive (mean 0.26) than L0 features (mean 0.34)---but the L17 endpoint is unambiguous. This is consistent with the progressive refinement hypothesis~\citep{kendiukhov2025circuits}: early and middle layers encode raw gene co-expression patterns (which, when amplified, maintain or disrupt progenitor-like states), while the final layers encode processed cell-identity representations whose amplification drives cells toward terminal differentiation.

This gradient mirrors the biological differentiation hierarchy~\citep{orkin2008hematopoiesis,graf2009forcing}, where early transcriptional programs maintain multipotency while late programs commit cells to specific fates. The finding that a transformer model spontaneously learns this hierarchical organization---without explicit supervision on differentiation stage---suggests that the layer-wise processing naturally decomposes cell state into a progression from raw features to commitment signals.

\subsection{Scale of the computational graph}

The 1,393,850 edges from a single source layer represent a lower bound on the model's circuit complexity. With 18 layers and $\sim$4,000 active features per layer, the complete circuit graph would contain on the order of tens of millions of edges. The signal attenuation pattern (L6: 695K $\to$ L11: 443K $\to$ L17: 256K) provides a natural compression---distant effects are weaker---but the sheer scale suggests that full circuit analysis of foundation models will require substantial computational investment and novel algorithmic approaches.

\subsection{Limitations}

Several limitations should be acknowledged. First, exhaustive tracing used only 20 cells per feature (vs.\ 200 in selective tracing), which reduces statistical power for detecting small effects and may explain the lower mean edges per feature (343 vs.\ 1,737 in selective tracing). Second, only 3 downstream layers were measured (L6, L11, L17) versus 12 in selective tracing, meaning intermediate-layer effects are not captured. Third, the three-way ablation tested only 8 triplets across 4 pathways; while results were consistent, broader coverage would strengthen the generalizability claim. Fourth, trajectory steering effects were small in absolute magnitude ($\sim$0.001--0.003), raising questions about biological significance despite directional consistency. Fifth, all experiments were conducted on Geneformer only; replication in other foundation models (scGPT~\citep{cui2024scgpt}, scBERT~\citep{yang2022scbert}) and in protein foundation models~\citep{rives2021biological} would test whether these architectural properties are model-specific or universal.

\subsection{Conclusions}

The transition from selective to exhaustive circuit analysis reveals that the ``map'' drawn by annotation-biased sampling systematically misrepresents the territory. The complete L5 circuit graph is dominated by unannotated hub features, organized in a heavy-tailed distribution, and characterized by deep redundancy without synergy. Combined with the causal evidence that late-layer (L17) features uniquely and universally drive cell state toward maturity while earlier layers do not, these results establish that Geneformer's internal representations encode a biologically meaningful functional specialization---implemented through massively redundant, hub-dominated circuits.

\section{Methods}

\subsection{Model and sparse autoencoders}

All experiments used Geneformer V2-316M~\citep{theodoris2023transfer}, an 18-layer, 18-head transformer model pre-trained on $\sim$100 million single-cell transcriptomes. TopK sparse autoencoders~\citep{makhzani2013ksparse} ($k = 32$, 4$\times$ overcomplete: $d_\text{model} = 1{,}152 \to d_\text{SAE} = 4{,}608$) were trained on residual stream~\citep{elhage2022superposition} activations at each layer using 4.05 million token positions from K562 cells, as described in \citet{kendiukhov2025sae_atlas}. Features were annotated by computing cosine similarity between decoder directions and gene embedding vectors, selecting the top 50 loading genes per feature, and performing enrichment analysis~\citep{subramanian2005gsea} against GO~\citep{ashburner2000go}, KEGG~\citep{kanehisa2000kegg}, and Reactome~\citep{jassal2020reactome} databases.

\subsection{Exhaustive circuit tracing}

For each of the 4,065 features at layer~5 with activation frequency $\geq 0.001$, we performed causal ablation tracing~\citep{pearl2009causality,vig2020causal,geiger2021causal} across 20 randomly sampled K562 cells. A clean forward pass cache was pre-computed once, storing source-layer hidden states, SAE encodings, and downstream clean SAE activations for all cells---a critical optimization that eliminated redundant forward passes, reducing computation from $\sim$20 days to 17.7 hours. For each feature, we set its TopK activation coefficient to zero in the source-layer residual stream, ran a forward pass from that layer, and encoded the downstream hidden states through the corresponding layer SAEs. Cohen's $d$ between clean and ablated activations was computed using Welford's online accumulator~\citep{welford1962note}. Edges with $|d| > 0.5$ and consistency $> 0.7$ were retained as significant. Three downstream layers (6, 11, 17) were measured. Total compute time was 17.7 hours on a single Apple M2 Max GPU (MPS backend).

\subsection{Higher-order combinatorial ablation}

Eight feature triplets were constructed across four biological pathways (vesicle transport: 3 triplets; mitosis: 2; metabolism: 2; cross-pathway DDR$\times$mitosis: 1), each containing features from layers 0, 5, and 11 with the same or related GO/pathway annotations. For each triplet, 7 ablation conditions were tested (A, B, C, AB, AC, BC, ABC) plus a clean baseline, with 200 K562 cells per condition. The three-way redundancy ratio was computed as:
\[
R_{ABC} = \frac{|d_{ABC}|}{|d_A| + |d_B| + |d_C|}
\]
The higher-order interaction term was computed via inclusion-exclusion:
\[
I_{ABC} = d_{ABC} - d_{AB} - d_{AC} - d_{BC} + d_A + d_B + d_C
\]
A target was classified as superadditive if $|d_{ABC}| > |d_A| + |d_B| + |d_C|$ (i.e., $R_{ABC} > 1$). Total compute time was 6.5 hours.

\subsection{Trajectory-guided feature steering}

Fourteen switch features identified in \citet{kendiukhov2025circuits} were selected for steering: 5 from L0, 2 from L5, 4 from L11, and 3 from L17. For each feature, we identified early-pseudotime cells (bottom 30\% of diffusion pseudotime~\citep{haghverdi2016diffusion}) from 481 Tabula Sapiens~\citep{tabula2022tabula} immune cells, where the feature was active. Steering was performed following the activation addition framework~\citep{turner2023activation,li2024inference}, by multiplying the feature's SAE activation coefficient by $\alpha \in \{2.0, 5.0\}$ and reconstructing the modified residual stream:
\[
h'_\ell = h_\ell + (\alpha - 1) \cdot a_f \cdot d_f
\]
where $a_f$ is the activation coefficient and $d_f$ is the decoder direction for feature $f$. The modified hidden state was then propagated through the remaining model layers to obtain logit changes.

State shift was quantified as:
\[
\Delta s = \cos(z', g_\text{late}) - \cos(z', g_\text{early}) - [\cos(z, g_\text{late}) - \cos(z, g_\text{early})]
\]
where $z$ and $z'$ are the clean and steered logit vectors, and $g_\text{late}$, $g_\text{early}$ are gene expression signatures computed from the top and bottom pseudotime deciles. Positive $\Delta s$ indicates a shift toward maturity. Total compute time was 2.1 minutes.

\subsection{Reproducibility}

All code is available at \url{https://github.com/Biodyn-AI/sae-biological-map}. Experiments were run on a MacBook Pro with Apple M2 Max (38-core GPU, 96~GB unified memory) using PyTorch 2.1 with MPS backend. Python environment: conda with NumPy 1.26.4, scanpy~\citep{wolf2018scanpy}. Total compute time for all experiments reported in this paper: $\sim$26.3 hours. Random seeds were fixed for reproducibility.

\section*{Data Availability}

The datasets analysed during the current study are available in the following repositories:
\begin{itemize}
\item \textbf{K562 CRISPRi perturbation data} (Replogle et al.~\citep{replogle2022mapping}): available at \url{https://plus.figshare.com/articles/dataset/Replogle_2022_K562_gwps/21452470} (Figshare, DOI: 10.6084/m9.figshare.21452470).
\item \textbf{Tabula Sapiens} single-cell atlas~\citep{tabula2022tabula}: available at \url{https://tabula-sapiens-portal.ds.czbiohub.org/} and via CZ CELLxGENE (\url{https://cellxgene.cziscience.com/collections/e5f58829-1a66-40b5-a624-9046778e74f5}).
\item \textbf{Geneformer V2-316M} pretrained model~\citep{theodoris2023transfer}: available on HuggingFace at \url{https://huggingface.co/ctheodoris/Geneformer} (subfolder \texttt{Geneformer-V2-316M}).
\item \textbf{Gene Ontology} annotations~\citep{ashburner2000go}: available at \url{http://geneontology.org/}.
\end{itemize}
All derived experimental data (trained SAE weights, feature catalogs, circuit graphs, and analysis outputs) are available at \url{https://github.com/Biodyn-AI/sae-biological-map}.

\section*{Code Availability}

All code for SAE training, feature analysis, causal patching, circuit tracing, and figure generation is available at \url{https://github.com/Biodyn-AI/sae-biological-map}.

\section*{Author Contributions}

I.K.\ conceived the study, designed and performed all experiments, analyzed the results, and wrote the manuscript.

\section*{Funding}

This work was self-funded by the author.

\section*{Competing Interests}

The author declares no competing interests.

\section*{Acknowledgments}

The author thanks the developers of Geneformer, scanpy, and PyTorch for making their software freely available.

\bibliographystyle{plainnat}
\bibliography{references_v3}

\end{document}